\documentclass[11pt]{article}
\usepackage{acl2014}
\usepackage{times}
\usepackage{url}
\usepackage{latexsym}
\usepackage{graphics,graphicx}
\usepackage{enumitem}
\usepackage{url}

\title{That's sick dude!: \\Automatic identification of word sense change across different timescales}

\author{Sunny Mitra${^1}$, Ritwik Mitra${^1}$, Martin Riedl${^2}$, \\
{\bf Chris Biemann${^2}$, Animesh Mukherjee${^1}$, Pawan Goyal${^1}$} \\
  ${^1}$Dept. of Computer Science and Engineering,\\ Indian Institute of Technology Kharagpur, India -- 721302 \\
  ${^2}$ FG Language Technology, Computer Science Department, TU Darmstadt, Germany\\
  ${^1}${\tt \{sunnym,ritwikm,animeshm,pawang\}@cse.iitkgp.ernet.in} \\
  ${^2}${\tt \{riedl,biem\}@cs.tu-darmstadt.de}
  }

\date{}

\begin{document}
\maketitle
\begin{abstract}

In this paper, we propose an unsupervised method to identify noun sense changes based on rigorous analysis of time-varying text data available in the form of millions of digitized books. We construct distributional thesauri based networks from data at different time points and cluster each of them separately to obtain word-centric sense clusters corresponding to the different time points. Subsequently, we compare these sense clusters of two different time points to find if (i) there is birth of a new sense or (ii) if an older sense has got split into more than one sense or (iii) if a newer sense has been formed from the joining of older senses or (iv) if a particular sense has died. We conduct a thorough evaluation of the proposed methodology both manually as well as through comparison with WordNet. Manual evaluation indicates that the algorithm could correctly identify 60.4\% birth cases from a set of 48 randomly picked samples and 57\% split/join cases from a set of 21 randomly picked samples. Remarkably, in 44\% cases the birth of a novel sense is attested by WordNet, while in 46\% cases and 43\% cases split and join are respectively confirmed by WordNet. Our approach can be applied for lexicography, as well as for applications like word sense disambiguation or semantic search.
 \end{abstract}

\section{Introduction}

Two of the fundamental components of a natural language communication are word sense discovery~\cite{Jones:86} and word sense disambiguation~\cite{Ide:98}. While discovery corresponds to acquisition of vocabulary, disambiguation forms the basis of understanding. These two aspects are not only important from the perspective of developing computer applications for natural languages but also form the key components of language evolution and change.

Words take different senses in different contexts while appearing with other words. Context plays a vital role in disambiguation of word senses as well as in the interpretation of the actual meaning of words. For instance, the word ``bank'' has several distinct interpretations, including that of a ``financial institution'' and the ``shore of a river.''  Automatic discovery and disambiguation of word senses from a given text is an important and challenging problem which has been extensively studied in the literature~\cite{Jones:86,Ide:98,Schutze:98,Navigli:09}. However, another equally important aspect that has not been so far well investigated corresponds to one or more changes that a word might undergo in its sense. This particular aspect is getting increasingly attainable as more and more time-varying text data become available in the form of millions of digitized books~\cite{Goldberg:13} gathered over the last centuries. As a motivating example one could consider the word ``sick'' -- while according to the standard English dictionaries the word is normally used to refer to some sort of illness, a new meaning of ``sick'' referring to something that is ``crazy'' or ``cool'' is currently getting popular in the English vernacular. This change is further interesting because while traditionally ``sick'' has been associated to something negative in general, the current meaning associates positivity with it. In fact, a rock band by the name of ``Sick Puppies'' has been founded which probably is inspired by the newer sense of the word sick. The title of this paper has been motivated by the above observation. Note that this phenomena of change in word senses has existed ever since the beginning of human communication~\cite{Bamman:11,Michel:11,Wijaya:11,Mihalcea:12}; however, with the advent of modern technology and the availability of huge volumes of time-varying data it now has become possible to automatically track such changes and, thereby, help the lexicographers in word sense discovery, and design engineers in enhancing various NLP/IR applications (e.g., disambiguation, semantic search etc.) that are naturally sensitive to change in word senses. 

The above motivation forms the basis of the central objective set in this paper, which is to devise a completely unsupervised approach to track noun sense changes in large texts available over multiple timescales. Toward this objective we make the following contributions: (a) devise a time-varying graph clustering based sense induction algorithm, (b) use the time-varying sense clusters to develop a split-join based approach for identifying new senses of a word, and (c) evaluate the performance of the algorithms on various datasets using different suitable approaches along with a detailed error analysis. Remarkably, comparison with the English WordNet indicates that in 44\% cases, as identified by our algorithm, there has been a birth of a completely novel sense, in 46\%  cases a new sense has split off from an older sense and in 43\% cases two or more older senses have merged in to form a new sense.  

The remainder of the paper is organized as follows. In the next section we present a short review of the literature. In Section~\ref{sec3} we briefly describe the datasets and outline the process of co-occurrence graph construction. In Section~\ref{sec4} we present an approach based on graph clustering to identify the time-varying sense clusters and in Section~\ref{sec5} we present the split-merge based approach for tracking word sense changes. Evaluation methods are summarized in Section~\ref{sec6}. Finally, conclusions and further research directions are outlined in Section~\ref{sec8}.

\section{Related work}\label{sec2}

Word sense disambiguation as well as word sense discovery have both remained key areas of research right from the very early initiatives in natural language processing research. Ide and Veronis~\shortcite{Ide:98} present a very concise survey of the history of ideas used in word sense disambiguation;  for a recent survey of the state-of-the-art one can refer to~\cite{Navigli:09}. Some of the first attempts to automatic word sense discovery were made by Karen Sp{\"a}rck Jones~\shortcite{Jones:86}; later in lexicography, it has been extensively used as a pre-processing step for preparing mono- and multi-lingual dictionaries~\cite{Kilgarriff:01,Kilgarriff:04}. However, as we have already pointed out that none of these works consider the temporal aspect of the problem.

In contrast, the current study, is inspired by works on language dynamics and opinion spreading~\cite{Mukherjee:11,Maity:12,Loreto:12} and automatic topic detection and tracking~\cite{Allan:98}. However, our work differs significantly from those proposed in the above studies. Opinion formation deals with the self-organisation and emergence of shared vocabularies whereas our work focuses on how the different senses of these vocabulary words change over time and thus become ``out-of-vocabulary''.  Topic detection involves detecting the occurrence of a new event such as a plane crash, a murder, a jury trial result, or a political scandal in a stream of news stories from multiple sources and tracking is the process of monitoring a stream of news stories to find those that track (or discuss) the same event. This is done on shorter timescales (hours, days), whereas our study focuses on larger timescales (decades, centuries) and we are interested in common nouns, verbs and adjectives as opposed to events that are characterized mostly by named entities. Other similar works on dynamic topic modelling can be found in~\cite{Blei:06,Wang:06}. Google books n-gram viewer\footnote{\url{ https://books.google.com/ngrams}} is a phrase-usage graphing tool which charts the yearly count of selected letter combinations, words, or phrases as found in over 5.2 million digitized books. It only reports frequency of word usage over the years, but does not give any correlation among them as e.g., in~\cite{Heyer:09}, and does not analyze their senses.

A few approaches suggested by~\cite{Bond:09,Paakko:12} attempt to augment WordNet synsets primarily using methods of annotation. Another recent work by Cook et al.~\shortcite{Cook:13} attempts to induce word senses and then identify novel senses by comparing two different corpora: the ``focus corpora'' (i.e., a recent version of the corpora) and the ``reference corpora'' (older version of the corpora). However, this method is limited as it only considers two time points to identify sense changes as opposed to our approach which is over a much larger timescale, thereby, effectively allowing us to track the points of change and the underlying causes. One of the closest work to what we present here has been put forward by~\cite{Tahmasebi:11}, where the authors analyze a newspaper corpus containing articles between 1785 and 1985. The authors mainly report the frequency patterns of certain words that they found to be candidates for change; however a detailed cause analysis as to why and how a particular word underwent a sense change has not been demonstrated. Further, systematic evaluation of the results obtained by the authors has not been provided.    

All the above points together motivated us to undertake the current work where we introduce, for the first time, a completely unsupervised and automatic method to identify the change of a word sense and the cause for the same. Further, we also present an extensive evaluation of the proposed algorithm in order to test its overall accuracy and performance. 

\section{Datasets and graph construction}\label{sec3}

In this section, we outline a brief description of the dataset used for our experiments and the graph construction procedure.  The primary source of data have been the millions of digitized books made available through the Google Book project~\cite{Goldberg:13}.
The Google Book syntactic n-grams dataset provides dependency fragment counts by the years. However, instead of using the plain syntactic n-grams, we use a far richer representation of the data in the form of a distributional thesaurus~\cite{Lin:97,Rychly:07}. In specific, we prepare a distributional thesaurus (DT) for each of the time periods separately and subsequently construct the required networks. We briefly outline the procedure of thesauri construction here referring the reader to~\cite{Riedl:13} for further details. In this approach, we first extract each word and a set of its context features, which are formed by labeled and directed dependency parse edges as provided in the dataset. Following this, we compute the frequencies of the word, the context and the words along with their context. Next we calculate the lexicographer's mutual information LMI~\cite{Kilgarriff:04} between a word and its features and retain only the top $1000$ ranked features for every word. Finally, we construct the DT network as follows: each word is a node in the network and the edge weight between two nodes is defined as the number of features that the two corresponding words share in common. 

\section{Tracking sense changes}\label{sec4}
The basic idea of our algorithm for tracking sense changes is as follows. If a word undergoes a sense change, this can be detected by comparing its senses obtained from two different time periods. Since we aim to detect this change automatically, we require distributional representations corresponding to word senses for different time periods. We, therefore, utilize the basic hypothesis of unsupervised sense induction to induce the sense clusters over various time periods and then compare these clusters to detect sense change. The basic premises of the `unsupervised sense induction' are briefly described below.

\subsection{Unsupervised sense induction}\label{subsec41}
 We use the co-occurrence based graph clustering framework introduced in~\cite{Biemann:06}. The algorithm proceeds in three basic steps. Firstly, a co-occurrence graph is created for every target word found in DT. Next, the neighbourhood/ego graph is clustered using the Chinese Whispers (CW) algorithm (see~\cite{Leskovec:12} for similar approaches). The algorithm, in particular, produces a set of clusters for each target word by decomposing its open neighborhood. We hypothesize that each different cluster corresponds to a particular sense of the target word. For a detailed description, the reader is referred to~\cite{Biemann:11}.

If a word undergoes sense change, this can be detected by comparing the sense clusters obtained from two different time periods by the algorithm outlined above. For this purpose, we use statistics from the DT corresponding to two different time intervals, say $tv_i$ and $tv_j$.  We then run the sense induction algorithm over these two different datasets. Now, for a given word $w$ that appears in both the datasets, we get two different set of clusters, say $C_i$ and $C_j$. Without loss of generality, let us assume that our algorithm detects $m$ sense clusters for the word $w$ in $tv_i$ and $n$ sense clusters in $tv_j$. Let $C_i=\{s_{i1},s_{i2},\ldots,s_{im}\}$ and $C_j=\{s_{j1},s_{j2},\ldots,s_{jn}\}$, where $s_{kz}$ denotes $z^{th}$ sense cluster for word $w$ during time interval $tv_k$. We next describe our algorithm for detecting sense change from these sets of sense clusters.

\subsection{Split, join, birth and death}\label{subsec42}
\begin{figure*}[hbt]
\centering
\includegraphics[width=14.4cm]{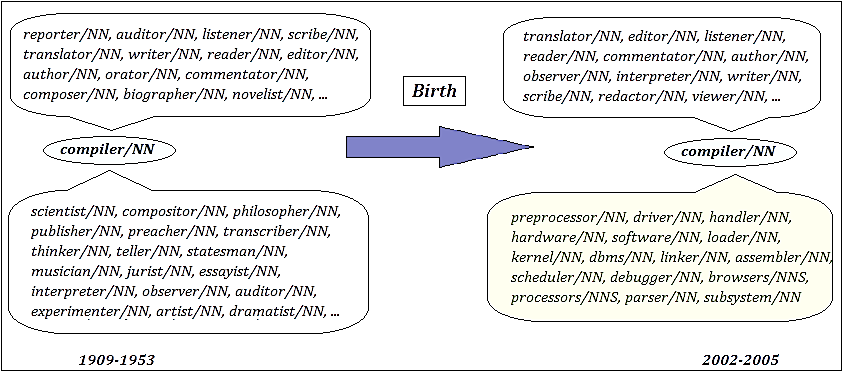} 
\caption{Example of the birth of a new sense for the word `compiler'} 
\label{figbirth}
\end{figure*}

We hypothesize that word $w$ can undergo sense change from one time interval ($tv_i$) to another ($tv_j$) as per one of the following scenarios:
\begin{description}[leftmargin=*]
\item[Split] A sense cluster $s_{iz}$ in $tv_i$ \textit{splits} into two (or more) sense clusters, $s_{jp_1}$ and $s_{jp_2}$ in $tv_j$
\item[Join] Two sense clusters $s_{iz_1}$ and $s_{iz_2}$ in $tv_i$ \textit{join} to make a single cluster $s_{jp}$ in $tv_j$
\item[Birth] A new sense cluster $s_{jp}$ appears in $tv_j$, which was absent in $tv_i$
\item[Death] A sense cluster $s_{iz}$ in $tv_i$ dies out and does not appear in $tv_j$
\end{description}
To detect split, join, birth or death, we build an $(m+1)\times (n+1)$ matrix $I$ to capture the intersection between sense clusters of two different time periods. The first $m$ rows and $n$ columns correspond to the sense clusters in $tv_i$ and $tv_j$ espectively. We append an additional row and column to capture the fraction of words, which did not show up in any of the sense clusters in another time interval. So, an element $I_{kl}$ of the matrix 
\begin{itemize}
\item $1\leq k\leq m, 1\leq l\leq n$: denotes the fraction of words in a newer sense cluster $s_{jl}$, that were also present in an older sense cluster $s_{ik}$. 
\item $k=m+1, 1\leq l\leq n$: denotes the fraction of words in the sense cluster $s_{jl}$, that were not present in any of the $m$ clusters in $tv_i$.
\item $1\leq k\leq m, l=n+1$: denotes the fraction of words in the sense cluster $s_{ik}$, that did not show up in any of the $n$ clusters in $tv_j$.
\end{itemize}
Thus, the matrix $I$ captures all the four possible scenarios for sense change. Since we can not expect a perfect split, birth etc., we used certain threshold values to detect if a candidate word is undergoing sense change via one of these four cases. In Figure~\ref{figbirth}, as an example, we illustrate the birth of a new sense for the word `compiler'.

\subsection{Multi-stage filtering}\label{subsec43}
To make sure that the candidate words obtained via our algorithm are meaningful, we applied multi-stage filtering to prune the candidate word list. The following criterion were used for the filtering:
\begin{description}[leftmargin=*]
\item[Stage 1] We utilize the fact that the CW algorithm is non-deterministic in nature. We apply CW three times over the source and target time intervals. We obtain the candidate word lists using our algorithm for the three runs, then take the intersection to output those words, which came up in all the three runs.
\item[Stage 2] From the above list, we retain only those candidate words, which have a part-of-speech tag `NN' or `NNS', as we focus on nouns for this work.
\item[Stage 3] We sort the candidate list obtained in Stage 2 as per their occurrence in the first time period. Then, we remove the top $20\%$ and the bottom $20\%$ words from this list. Therefore, we consider the {\em torso} of the frequency distribution which is the most informative part for this type of an analysis.
\end{description}

\section{Experimental framework}\label{sec5}
For our experiments, we utilized DTs created for 8 different time periods: 1520-1908, 1909-1953, 1954-1972, 1973-1986, 1987-1995, 1996-2001, 2002-2005 and 2006-2008~\cite{Biemann:14}. The time periods were set such that the amount of data in each time period is roughly the same. We will also use $T_1$ to $T_8$ to denote these time periods. The parameters for CW clustering were set as follows. The size of the neighbourhood ($N$) to be clustered was set to $200$. The parameter $n$ regulating the edge density in this neighbourhood was set to $200$ as well. The parameter $a$ was set to $lin$, which corresponds to favouring smaller clusters by hub downweighing\footnote{data available at \url{http://sf.net/p/jobimtext/wiki/LREC2014_Google_DT/}}. The threshold values used to detect the sense changes were as follows. For birth, at least $80\%$ words of the target cluster should be novel. For split, each split cluster should have at least $30\%$ words of the source cluster and the total intersection of all the split clusters should be $>80\%$. The same parameters were used for the join and death case with the interchange of source and target clusters.

\subsection{Signals of sense change}
Making comparisons between all the pairs of time periods gave us 28 candidate words lists. For each of these comparison, we applied the multi-stage filtering to obtain the pruned list of candidate words. Table \ref{tab:candidates} provides some statistics about the number of candidate words obtained corresponding to the birth case. The rows correspond to the source time-period and the columns correspond to the target time periods. An element of the table shows the number of candidate words obtained by comparing the corresponding source and target time periods.

\begin{table}[h]
\begin{center}
\caption{Number of candidate birth senses between all time periods}
\resizebox{7.7cm}{!}{
\begin{tabular}{|l|r|r|r|r|r|r|r|r|}\hline
      &$T_2$&$T_3$&$T_4$&$T_5$&$T_6$&$T_7$&$T_8$\\\hline
$T_1$&2498&3319&3901&4220&4238&4092&3578\\\hline
$T_2$&        &1451&2330&2789&2834&2789&2468\\\hline
$T_3$&        &        & 917 &1460&1660&1827&1815\\\hline
$T_4$&        &        &        &  517& 769&1099&1416\\\hline
$T_5$&        &        &        &        & 401& 818&1243\\\hline
$T_6$&        &        &        &        &        & 682&1107\\\hline
$T_7$&        &        &        &        &        &       &609\\\hline
\end{tabular}
}
\label{tab:candidates}
\end{center}
\end{table}
The table clearly shows a trend. For most of the cases, the number of candidate birth senses tends to increase as we go from left to right. Similarly, this number decreases as we go down in the table. This is quite intuitive since going from left to right corresponds to increasing the gap between two time periods while going down corresponds to decreasing this gap. As the gap increases (decreases), one would expect more (less) new senses coming in. Even while moving diagonally, the candidate words tend to decrease as we move downwards. This corresponds to the fact that the number of years in the time periods decreases as we move downwards, and therefore, the gap also decreases.

\subsection{Stability analysis \& sense change location}
Formally, we consider a sense change from $tv_i$ to $tv_j$ \textbf{stable} if it was also detected while comparing $tv_i$ with the following time periods $tv_k$s. This number of subsequent time periods, where the same sense change is detected, helps us to determine the \textbf{age} of a new sense. Similarly, for a candidate sense change from $tv_i$ to $tv_j$, we say that the \textbf{location} of the sense change is $tv_j$ if and only if that sense change does not get detected by comparing $tv_i$ with any time interval $tv_k$, intermediate between $tv_i$ and $tv_j$. 

Table \ref{tab:candidates} gives a lot of candidate words for sense change. However, not all the candidate words were stable. Thus, it was important to prune these results using stability analysis. Also, it is to be noted that these results do not pin-point to the exact time-period, when the sense change might have taken place. For instance, among the $4238$ candidate birth sense detected by comparing $T_1$ and $T_6$, many of these new senses might have come up in between $T_2$ to $T_5$ as well. We prune these lists further based on the stability of the sense, as well as to locate the approximate time interval, in which the sense change might have occurred. 

Table \ref{tab:stab} shows the number of stable (at least twice) senses as well as the number of stable sense changes located in that particular time period. While this decreases recall, we found this to be beneficial for the accuracy of the method. 

\begin{table}[h]
\begin{center}
\caption{Number of candidate birth senses obtained for different time periods}
\resizebox{7.7cm}{!}{
\begin{tabular}{|l|r|r|r|r|r|r|r|r|}\hline
      &$T_2$&$T_3$&$T_4$&$T_5$&$T_6$&$T_7$\\\hline
$T_1$&2498&3319&3901&4220&4238&4092\\
stable&537&989&1368&1627&1540&1299\\
located&537&754&772&686&420&300\\\hline
$T_2$&        &1451&2330&2789&2834&2789\\
stable&        &343&718&938&963&810\\
located&      &343&561&517&357&227\\\hline
\end{tabular}
}
\label{tab:stab}
\end{center}
\end{table}

\begin{figure*}[hbt]
\centering
\includegraphics[width=0.92\textwidth]{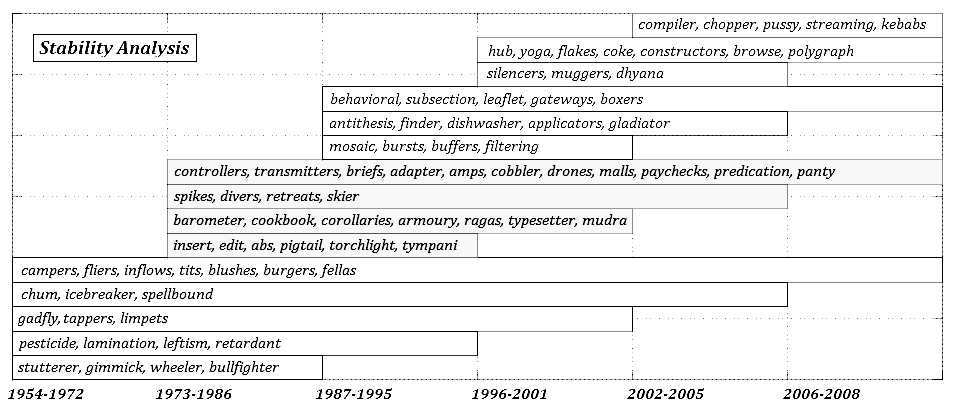} 
\caption{Examples of birth senses placed on a timeline as per their location as well as age} 
\label{fig:stab}
\end{figure*}

Once we were able to locate the senses as well as to find the age of the senses, we attempted to select some representative words and plotted them on a timeline as per the birth period and their age in Figure~\ref{fig:stab}. The source time period here is 1909-1953. 

\section{Evaluation framework}\label{sec6}

During evaluation, we considered the clusters obtained using the 1909-1953 time-slice as our reference and attempted to track sense change by comparing these with the clusters obtained for 2002-2005. The sense change detected was categorized as to whether it was a new sense (birth), a single sense got split into two or more senses (split) or two or more senses got merged (join) or a particular sense died (death). We present a few instances of the resulting clusters in the paper and refer the reader to the supplementary material\footnote{\url{http://cse.iitkgp.ac.in/resgrp/cnerg/acl2014_wordsense/}} for the rest of the results. 

\subsection{Manual evaluation}
The algorithm detected a lot of candidate words for the cases of birth, split/join as well as death. Since it was difficult to go through all the candidate sense changes for all the comparisons manually, we decided to randomly select some candidate words, which were flagged by our algorithm as undergoing sense change, while comparing 1909-1953 and 2002-2005 DT. We selected 48 random samples of candidate words for birth cases and 21 random samples for split/join cases. One of the authors annotated each of the birth cases identifying whether or not the algorithm signalled a true sense change while another author did the same task for the split/join cases. The accuracy as per manual evaluation was found to be 60.4\% for the birth cases and 57\% for the split/join cases.

Table \ref{tab:birthEval} shows the evaluation results for a few candidate words, flagged due to birth. Columns correspond to the candidate words, words obtained in the cluster of each candidate word (we will use the term `birth cluster' for these words, henceforth), which indicated a new sense, the results of manual evaluation as well as the possible sense this birth cluster denotes. 

\begin{table*}[!thb]
\begin{center}
\caption{Manual evaluation for seven randomly chosen candidate birth clusters between time periods 1909-1953 and 2002-2005 \\ }
\resizebox{14.2cm}{!}{
\begin{tabular}{|l|l|l|l|}\hline
\bf{Sl} & \bf{Candidate} & \bf{birth cluster} & \bf{Evaluation judgement, }\\
\bf{No.} & \bf{Word} &  & \bf{comments}  \\\hline
1 & implant & {\sl gel, fibre, coatings, cement, materials, metal, filler} & \textbf{No}, New set of words but \\
   &              & {\sl silicone, composite, titanium, polymer, coating} & similar sense already existed\\\hline
2 & passwords & {\sl browsers, server, functionality, clients, workstation} & \textbf{Yes}, New sense related \\
   &                  & {\sl printers, software, protocols, hosts, settings, utilities} & to `a computer sense'\\\hline
3 & giants & {\sl multinationals, conglomerates, manufacturers} & \textbf{Yes}, New sense as `an \\
     &           & {\sl corporations, competitors, enterprises, companies} & organization with very great \\
     &           & {\sl businesses, brands, firms} & size or force'\\\hline
4 & donation & {\sl transplantation, donation, fertilization, transfusions} & \textbf{Yes}, The new usage of donation \\
     &                & {\sl transplant, transplants, insemination, donors, donor ...} & associated with body organs etc. \\\hline
5 & novice & {\sl negro, fellow, emigre, yankee, realist, quaker, teen} & \textbf{No},  this looks like a false\\
     &            & {\sl male, zen, lady, admiring, celebrity, thai, millionaire ...} &  positive\\\hline
6 & partitions & {\sl server, printers, workstation, platforms, arrays} & \textbf{Yes}, New usage related to \\
     &                 & {\sl modules, computers, workstations, kernel ...} & the `computing' domain\\\hline
7 & yankees & {\sl athletics, cubs, tigers, sox, bears, braves, pirates} & \textbf{Yes}, related to the `New \\
     &               & {\sl cardinals, dodgers, yankees, giants, cardinals ...} & York Yankees' team\\\hline
\end{tabular}
}
\label{tab:birthEval}
\end{center}
\end{table*}

Table \ref{tab:splitEval} shows the corresponding evaluation results for a few candidate words, flagged due to split or join. 

\begin{table*}[!thb]
\begin{center}
\caption{Manual evaluation for five randomly chosen candidate split/join clusters between time periods 1909-1953 and 2002-2005 \\}
\resizebox{14cm}{!}{
\begin{tabular}{|l|l|l|}\hline
\bf{Sl} & \bf{Candidate} & \bf{Source and target clusters}\\
\bf{No.} & \bf{Word} &  \\\hline
1 & intonation & $S$: {\sl whisper, glance, idioms, gesture, chant, sob, inflection, diction, sneer, rhythm, accents ...}\\
   & (\bf{split}) & $T_1$: {\sl nod, tone, grimace, finality, gestures, twang, shake, shrug, irony, scowl, twinkle ...}\\
   &                  & $T_2$: {\sl accents, phrase, rhythm, style, phonology, diction, utterance, cadence, harmonies ...}\\\hline
   &\multicolumn{2}{|l|}{\textbf{Yes}, $T_1$ corresponds to intonation in normal conversations while $T_2$ corresponds to the use of accents in}\\
  &\multicolumn{2}{|l|}{formal and research literature}\\\hline
2 & diagonal & $S$: {\sl coast, edge, shoreline, coastline, border, surface, crease, edges, slope, sides, seaboard ...}\\
   & (\bf{split}) & $T_1$: {\sl circumference, center, slant, vertex, grid, clavicle, margin, perimeter, row, boundary ..}\\
   &                  & $T_2$: {\sl border, coast, seaboard, seashore, shoreline, waterfront, shore, shores, coastline, coasts}\\\hline
   &\multicolumn{2}{|l|}{\textbf{Yes}, the split $T_1$ is based on mathematics where as $T_2$ is based on geography}\\\hline
3 & mantra & $S_1$: {\sl sutra, stanza, chanting, chants, commandments, monologue, litany, verse, verses ...}\\
   & (\bf{join})     & $S_2$: {\sl praise, imprecation, benediction, praises, curse, salutation, benedictions, eulogy ...}\\
   &                    & $T$: {\sl blessings, spell, curses, spells, rosary, prayers, blessing, prayer, benediction ...}\\\hline
   &\multicolumn{2}{|l|}{\textbf{Yes}, the two seemingly distinct senses of mantra - a contextual usage for chanting and prayer ($S_1$)}\\
   &\multicolumn{2}{|l|}{and another usage in its effect - salutations, benedictions ($S_2$) have now merged in $T$.}\\\hline
4 & continuum & $S$: {\sl circumference, ordinate, abscissa, coasts, axis, path, perimeter, arc, plane  axis ...}\\
   & (\bf{split}) & $T_1$: {\sl  roadsides, corridors, frontier, trajectories, coast, shore, trail, escarpment, highways ...}\\
   &                  & $T_2$: {\sl arc, ellipse, meridians, equator, axis, axis, plane, abscissa, ordinate, axis, meridian ....}\\\hline
   &\multicolumn{2}{|l|}{\textbf{Yes}, the split $S_1$ denotes the usage of `continuum' with physical objects while the}\\
   &\multicolumn{2}{|l|}{the split $S_2$ corresponds to its usages in mathematics domain.}\\\hline
5 & headmaster & $S_1$: {\sl master, overseer, councillor, chancellor, tutors, captain, general, principal ...}\\
   & (\bf{join})     & $S_2$: {\sl mentor, confessor, tutor, founder, rector, vicar, graduate, counselor, lawyer ...}\\
   &                    & $T$: {\sl chaplain, commander, surveyor, coordinator, consultant, lecturer, inspector ...}\\\hline
   &\multicolumn{2}{|l|}{\textbf{No}, it seems a false positive}\\\hline
\end{tabular}
}
\label{tab:splitEval}
\end{center}
\end{table*}

A further analysis of the words marked due to birth in the random samples indicates that there are  22 technology-related words, 2 slangs, 3 economics related words and 2 general words.  For the split-join case we found that there are 3 technology-related words while the rest of the words are general. Therefore one of the key observations is that most of the technology related words (where the neighborhood is completely new) could be extracted from our birth results. In contrast, for the split-join instances most of the results are from the general category since the neighborhood did not change much here; it either got split or merged from what it was earlier.

\subsection{Automated evaluation with WordNet}
In addition to manual evaluation, we also performed automated evaluation for the candidate words. We chose WordNet for automated evaluation because not only does it have a wide coverage of word senses but also it is being maintained and updated regularly to incorporate new senses. We did this evaluation for the candidate birth, join and split sense clusters obtained by comparing 1909-1953 time period with respect to 2002-2005. For our evaluation, we developed an aligner to align the word clusters obtained with WordNet senses. The aligner constructs a WordNet dictionary for the purpose of synset alignment. The CW cluster is then aligned to WordNet synsets by comparing the clusters with WordNet graph and the synset with the maximum alignment score is returned as the output. In summary, the aligner tool takes as input the CW cluster and returns a WordNet synset id that corresponds to the cluster words. The evaluation settings were as follows:

\begin{description}[leftmargin=*]
\item[Birth:] For a candidate word flagged as birth, we first find out the set of all WordNet synset ids for its CW clusters in the source time period (1909-1953 in this case). Let $S_{init}$ denote the union of these synset ids. We then find WordNet synset id for its birth-cluster, say $s_{new}$. Then, if $s_{new}\notin S_{init}$, it implies that this is a new sense that was not present in the source clusters and we call it a `success' as per WordNet. 
\item[Join:] For the join case, we find WordNet synset ids $s_1$ and $s_2$ for the clusters obtained in the source time period and $s_{new}$ for the join cluster in the target time period. If $s_1\neq s_2$ and $s_{new}$ is either $s_1$ or $s_2$, we call it a `success'.
\item[Split:] For the split case, we find WordNet synset id $s_{old}$ for the source cluster and synset ids $s_{1}$ and $s_{2}$ for the target split clusters. If $s_1\neq s_2$ and either $s_1$, or $s_2$ retains the id $s_{old}$, we call it a `success'.
\end{description}

Table \ref{tab:wordEval} show the results of WordNet based evaluation. In case of birth we observe a success of 44\% while for split and join we observe a success of 46\% and 43\% respectively. 
\begin{table}
\begin{center}
\caption{Results of the automatic evaluation using WordNet}
\resizebox{7.7cm}{!}{
\begin{tabular}{|l|l|l|l|}\hline
\bf{Category} & \bf{No. of Candidate Words} & \bf{Success Cases}\\\hline
Birth & 810 & 44$\%$\\
Split & 24 & 46$\%$\\
Join & 28 & 43$\%$\\\hline
\end{tabular}
}
\label{tab:wordEval}
\end{center}
\end{table}
We then manually verified some of the words that were deemed as successes, as well as investigated WordNet sense they were mapped to. Table \ref{tab:wordEval1} shows some of the words for which the evaluation detected success along with WordNet senses. Clearly, the cluster words correspond to a newer sense for these words and the mapped WordNet synset matches the birth cluster to a very high degree.

\begin{table*}[thb]
\begin{center}
\caption{Example of randomly chosen candidate birth clusters mapped to WordNet}
\resizebox{14.2cm}{!}{
\begin{tabular}{|l|l|l|l|}\hline
\bf{Sl} & \bf{Candidate} & \bf{birth cluster} & \bf{Synset Id, }\\
\bf{No.} & \bf{Word} &  & \bf{WordNet sense}  \\\hline
1 & macro & {\sl code, query, handler, program, procedure, subroutine} & \textbf{6582403}, a set sequence of steps, \\
   &              & {\sl module, script} & part of larger computer program\\\hline
2 & caller & {\sl browser, compiler, sender, routers, workstation, cpu} & \textbf{4175147}, a computer that \\
   &                  & {\sl host, modem, router, server} & provides client stations with access to files\\\hline
3 & searching & {\sl coding, processing, learning, computing, scheduling}& \textbf{1144355}, programming: setting an \\
     &           & {\sl planning, retrieval, routing, networking, navigation} & order and time for planned events  \\\hline
4 & hooker & {\sl bitch, whore, stripper, woman slut, prostitute} & \textbf{10485440}, a woman who \\
     &                & {\sl girl, dancer ...} & engages in sexual intercourse for money  \\\hline
5 & drones & {\sl helicopters, fighters, rockets, flights, planes} & \textbf{4264914}, a craft capable of \\
     &            & {\sl vehicles, bomber, missions, submarines ...} &  traveling in outer space\\\hline
6 & amps & {\sl inverters, capacitor, oscillators, switches, mixer} & \textbf{2955247}, electrical device characterized \\
     &                 & {\sl transformer, windings, capacitors, circuits ...} & by its capacity to store an electric charge\\\hline
7 & compilers & {\sl interfaces, algorithms, programming, software} & \textbf{6566077}, written programs pertaining \\
     &           & {\sl modules, libraries, routines, tools, utilities ...} & to the operation of a computer system \\\hline
\end{tabular}
}
\label{tab:wordEval1}
\end{center}
\end{table*}
\begin{table*}[!thb]
\begin{center}
\caption{Some representative examples for candidate death sense clusters}
\resizebox{14.2cm}{!}{
\begin{tabular}{|l|l|l|l|}\hline
\bf{Sl} & \bf{Candidate} & \bf{death cluster} & \bf{Vanished meaning}\\
\bf{No.} & \bf{Word} &  & \\\hline
1 & slop & {\sl jeans, velveteen, tweed, woollen, rubber, sealskin, wear} & clothes and bedding supplied to \\
   &         & {\sl oilskin, sheepskin, velvet, calico, deerskin, goatskin, cloth ...}  & sailors by the navy\\\hline
2 & blackmail & {\sl subsidy, rent, presents, tributes, money, fine, bribes} & Origin: denoting protection money \\
   &                 & {\sl dues, tolls, contributions, contribution, customs, duties ...} & levied by Scottish chiefs \\\hline
3 & repertory & {\sl dictionary, study, compendium, bibliography, lore, directory} & Origin: denoting an index \\
    &                & {\sl catalogues, science, catalog, annals, digest, literature ...} & or catalog: from late Latin repertorium\\\hline
4 & phrasing & {\sl contour, outline, construction, handling, grouping, arrangement} & in the sense `style or manner of \\
    &               & {\sl structure, modelling, selection, form ...} & expression': via late Latin Greek phrasis\\\hline
\end{tabular}
}
\label{tab:vanishEx}
\end{center}
\end{table*}

\subsection{Evaluation with a slang list}
Slangs are words and phrases that are regarded as very informal, and are typically restricted to a particular context. New slang words come up every now and then, and this plays an integral part in the phenomena of sense change. We therefore decided to perform an evaluation as to how many slang words were being detected by our candidate birth clusters. We used a list of slangs available from the slangcity website\footnote{\url{http://slangcity.com/email_archive/index_2003.htm}}. We collected slangs for the years 2002-2005 and found the intersection with our candidate birth words. Note that the website had a large number of multi-word expressions that we did not consider in our study. Further, some of the words appeared as either erroneous or very transient (not existing more than a few months) entires, which had to be removed from the list. All these removal left us with a very little space for comparison; however, despite this we found 25 slangs from the website that were present in our birth results, e.g. `bum', `sissy', `thug', `dude' etc.

\subsection{Evaluation of candidate death clusters}
Much of our evaluation was focussed on the birth sense clusters, mainly because these are more interesting from a lexicographic perspective. Additionally, the main theme of this work was to detect new senses for a given word. To detect a true death of a sense, persistence analysis was required, that is, to verify if the sense was persisting earlier and vanished after a certain time period. While such an analysis goes beyond the scope of this paper, we selected some interesting candidate ``death" senses. Table \ref{tab:vanishEx} shows some of these interesting candidate words, their death cluster along with the possible vanished meaning, identified by the authors. While these words are still used in a related sense, the original meaning does not exist in the modern usage.

\section{Conclusions}\label{sec8}

In this paper, we presented a completely unsupervised method to detect word sense changes by analyzing millions of digitized books archived spanning several centuries. In particular, we constructed DT networks over eight different time windows, clustered these networks and compared these clusters to identify the emergence of novel senses. The performance of our method has been evaluated manually as well as by comparison with WordNet and a list of slang words. Through manual evaluation we found that the algorithm could correctly identify 60.4\% birth cases from a set of 48 random samples and 57\% split/join cases from a set of 21 randomly picked samples. Quite strikingly, we observe that (i) in 44\% cases the birth of a novel sense is attested by WordNet, (ii) in 46\% cases the split of an older sense is signalled on comparison with WordNet and (iii) in 43\% cases the join of two senses is attested by WordNet. These results might have strong lexicographic implications -- even if one goes by very moderate estimates almost half of the words would be candidate entries in WordNet if they were not already part of it. This method can be extremely useful in the construction of lexico-semantic networks for low-resource languages, as well as for keeping lexico-semantic resources up to date in general. 

Future research directions based on this work are manifold. On one hand, our method can be used by lexicographers in designing new dictionaries where candidate new senses can be semi-automatically detected and included, thus greatly reducing the otherwise required manual effort. On the other hand, this method can be directly used for various NLP/IR applications like semantic search, automatic word sense discovery as well as disambiguation. For semantic search, taking into account the newer senses of the word can increase the relevance of the query result. Similarly, a disambiguation engine informed with the newer senses of a word can increase the efficiency of disambiguation, and recognize senses uncovered by the inventory that would otherwise have to be wrongly assigned to covered senses. In addition, this method can be also extended to the `NNP' part-of-speech (i.e., named entities) to identify changes in role of a person/place. Furthermore, it would be interesting to apply this method to languages other than English and to try to align new senses of cognates across languages.   

\section*{Acknowledgements}

AM would like to thank DAAD for supporting the faculty exchange programme to TU Darmstadt. PG would like to thank Google India Private Ltd. for extending travel support to attend the conference. MR and CB have been supported by an IBM SUR award and by LOEWE as part of the research center {\em Digital Humanities}.


\end{document}